%% file: main.tex
\pdfoutput=1
\documentclass[11pt]{article}

\usepackage[final]{acl}

\usepackage{times}
\usepackage{latexsym}

\usepackage[T1]{fontenc}
\usepackage[utf8]{inputenc}

\usepackage{microtype}

\usepackage{inconsolata}
\usepackage{caption} 
\usepackage{diagbox}
\usepackage{multirow}
\usepackage{subcaption}
\usepackage{mdframed}
\usepackage{adjustbox}
\usepackage[most]{tcolorbox}
\tcbuselibrary{breakable, skins}
\include{math_command}

\include{custom_command}
\usepackage{cleveref}

\NewTColorBox{NewBox}{ s O{!htbp} m m }{%
  floatplacement={#2},
  IfBooleanTF={#1}{float*,width=\textwidth}{float},
  colframe=green!50!black,colback=green!10!white, % any tcolorbox options here
  after upper={\captionof{figure}{#3}\label{#4}}, % Caption and label
}

\title{Sequence-level Large Language Model Training with Contrastive Preference Optimization}

\author{Zhili Feng \\
  Carnegie Mellon University\thanks{Work done as an intern at Amazon.} \\
  \texttt{zhilif@andrew.cmu.edu} \\\And
  Dhananjay Ram \\
  AWS AI \\ \texttt{radhna@amazon.com} \\\And
  Cole Hawkins \\
  AGI Foundations\\ \texttt{colehawk@amazon.com} \\\AND
  Aditya Rawal \\
  AGI Foundations\\ \texttt{adirawal@amazon.com} \\\And
  Jinman Zhao \\
  AGI Foundations\\ \texttt{jinming@amazon.com} \\\And
  Sheng Zha \\
  AGI Foundations\\ \texttt{zhasheng@amazon.com}
  }

\begin{document}
\maketitle
\begin{abstract}
The next token prediction loss is the dominant self-supervised training objective for large language models and has achieved promising results in a variety of downstream tasks. However, upon closer investigation of this objective, we find that it lacks an understanding of sequence-level signals, leading to a mismatch between training and inference processes. To bridge this gap, we introduce a contrastive preference optimization (CPO) procedure that can inject sequence-level information into the language model at any training stage without expensive human labeled data. Our experiments show that the proposed objective surpasses the next token prediction in terms of win rate in the instruction-following and text generation tasks. 
%Specifically, using OpenLlama-3B, our method achieves a $13.8\%$ improvement in an instruction-following task and a $3\%$ increase in a text-generation task.
\end{abstract}

\input{intro}

\input{related_work}
% \input{prelim}
\input{cpo}

\input{experiment}

\section{Conclusions and Limitations}
In this paper, we propose an auxiliary CPO loss function for LLM training, which can be used with or without ranking signals, depending on the quality of the negative samples. We investigated several ways to generate negative samples. One limitation of this work is that the synthetic data are very noisy unless generated autoregressively; it is interesting to explore other ways to efficiently generate high-quality negative data beyond the autoregressive fashion. 
%One possible direction is to consider Langevin dynamic sampling, which samples all tokens in parallel. 

% \bibliographystyle{acl_natbib}
\bibliography{custom_citation.bib}
\appendix
\input{appendix}

\end{document}

%% file: math_command.tex
\usepackage{amsmath,amsfonts,bm}

\def\eqref#1{equation~\ref{#1}}
\def\1{\bm{1}}

\def\ve{{\bm{e}}}

\def\vx{{\bm{x}}}
\def\vy{{\bm{y}}}

\DeclareMathAlphabet{\mathsfit}{\encodingdefault}{\sfdefault}{m}{sl}
\SetMathAlphabet{\mathsfit}{bold}{\encodingdefault}{\sfdefault}{bx}{n}

\def\gD{{\mathcal{D}}}

\def\gX{{\mathcal{X}}}

\newcommand{\E}{\mathbb{E}}

\newcommand{\R}{\mathbb{R}}

\newcommand{\KL}{D_{\mathrm{KL}}}

%% file: custom_command.tex
\usepackage{mathtools}
\usepackage{amsmath,amsfonts,amssymb,amsthm,commath}

\usepackage{graphicx}
\graphicspath{ {./figures/} }
\usepackage{caption}
\usepackage{multirow}
\usepackage{booktabs}
\usepackage{subcaption}
\usepackage{listings}
\usepackage{tikz,forest}
\usetikzlibrary{arrows.meta}
\usetikzlibrary{trees}
\usetikzlibrary{er,positioning}
\usetikzlibrary{fit,positioning}
\tikzstyle{block} = [rectangle, draw, fill=blue!20, 
    text width=12.8em, text centered, rounded corners, minimum height=4em]
\tikzstyle{line} = [draw, -latex']
\tikzstyle{cloud} = [draw, ellipse,fill=red!20, node distance=3cm,
    minimum height=2em]

\forestset{
    .style={
        for tree={
            base=bottom,
            child anchor=north,
            align=center,
            s sep+=1cm,
    straight edge/.style={
        edge path={\noexpand\path[\forestoption{edge},thick,-{Latex}] 
        (!u.parent anchor) -- (.child anchor);}
    },
    if n children={0}
        {tier=word, draw, thick, rectangle}
        {draw, diamond, thick, aspect=2},
    if n=1{%
        edge path={\noexpand\path[\forestoption{edge},thick,-{Latex}] 
        (!u.parent anchor) -| (.child anchor) node[pos=.2, above] {Y};}
        }{
        edge path={\noexpand\path[\forestoption{edge},thick,-{Latex}] 
        (!u.parent anchor) -| (.child anchor) node[pos=.2, above] {N};}
        }
        }
    }
}

\usepackage[inline]{enumitem}

\def\ddefloop#1{\ifx\ddefloop#1\else\ddef{#1}\expandafter\ddefloop\fi}
\def\ddef#1{\expandafter\def\csname bb#1\endcsname{\ensuremath{\mathbb{#1}}}}
\ddefloop ABCDEFGHIJKLMNOPQRSTUVWXYZ\ddefloop
\def\ddef#1{\expandafter\def\csname c#1\endcsname{\ensuremath{\mathcal{#1}}}}
\ddefloop ABCDEFGHIJKLMNOPQRSTUVWXYZ\ddefloop

\usepackage{tikz}

\def\pr{\pi_{\operatorname{ref}}}

\newcommand*\subtxt[1]{_{\textnormal{#1}}}
\DeclareRobustCommand\_{\ifmmode\expandafter\subtxt\else\textunderscore\fi}

\DeclarePairedDelimiter\autobracket{(}{)}
\newcommand{\p}[1]{\autobracket*{#1}}

\newcommand{\ip}[2]{\left\langle #1, #2 \right \rangle}
\theoremstyle{definition}

%% file: intro.tex
\section{Introduction}

Next token prediction is now the predominant way for pre-training and supervised fine-tuning (SFT) of large language models (LLM). This loss function can be easily scaled up to train models with trillions of parameters and tokens, and it has demonstrated the ability to generate coherent and contextually relevant text. Let $P$ be the unknown target language distribution and let $Q$ be the distribution of our model at hand. The goal of next token prediction is to minimize the \textit{forward-KL} divergence between $P$ and $Q$. This training process only supervises the prediction of one token at a time, given the full context of the ground truth. On the other hand, during inference, the model needs to generate a whole sequence (for a given prompt) relying on its own prior predictions. This mismatch between the training and inference stage is known as \textit{exposure-bias} in the literature of RNN and sequence-to-sequence model \citep{bengio2015scheduled,ranzato2015sequence}.

In other words, next token prediction injects only \textit{token-level} information into the model, but missing \textit{sequence-level} signal. The latter requires a generation of a longer horizon, which often relies on reinforcement learning algorithms; for example, reinforcement learning with human feedback (RLHF) \citep{ouyang2022training}; and is computationally expensive.
In this work, we ask the following question:
	\textit{Can we introduce sequence-level information in LLM pre-training / SFT with a small computational cost?}

We answer the question affirmatively with our proposed \textsc{\textbf{C}ontrastive \textbf{P}reference \textbf{O}ptimization} (CPO) method. The goal of CPO is to improve \textbf{generation quality}. Unlike RLHF, the proposed CPO method does not require human preference information as the training signal. While we demonstrate CPO in the SFT case, the loss can be seemlessly applied to the late stage of pretraining as well.

%Another related method that optimizes the quality of generated text is BRIO \citep{liu2022brio}. Although unlike BRIO, the proposed CPO method does not necessarily rely on autoregressively sampled negative sequences from the model and therefore is much more computational efficient and easier to scale up. In addition, CPO is also derived from a more principled statistical perspective. Our experiments demonstrate that CPO is able to improve the quality of text generation in terms of reward model scores and reverse-KL divergence.

%% file: related_work.tex
\section{Related work}

LLMs trained with next token prediction loss \citep{radford2019language,chung2022scaling,sanh2021multitask,zhou2023lima} have demonstrated many fascinating capabilities, including the ability to perform zero-shot or few-shot tasks \citep{radford2019language,brown2020language} and the ability to reason \citep{wei2022chain}. 

Several works have investigated the shortcomings of MLE and exposure bias. \citet{arora2022exposure} measured the accumulation of errors in language generation due to exposure bias. \citet{schmidt2019generalization} connected exposure bias to generalization. \citet{wang2020exposure} studied how exposure bias leads to hallucination in neural machine translation. To mitigate exposure bias, there exists a long line of work that has explored sequence-level training methods. \citet{bengio2015scheduled,ranzato2015sequence} proposed to train RNN with RL or RL-related algorithms rather than teacher-forcing. BRIO \citet{liu2022brio} targeted the summarization task with the ROUGE signal. \citet{pang2020text} trained the language models with an offline RL algorithm. There also exists a line of works that generate samples during training and mix the samples with ground truth data \citep{shen2015minimum,zhang2019bridging,duckworth2019parallel}. 

Recently, RLHF \citep{stiennon2020learning,ouyang2022training} and its supervised version DPO \citep{rafailov2023direct} were developed for alignment. They are effectively sequence-level training techniques. These algorithms require a pair of preferred and rejected samples, which are usually gathered by human labeling. The RL approach to language modeling is also closely related to energy-based models (EBM) \citep{korbak2022rl, deng2020residual}. This EBM form has also been studied in controlled text generation \citet{kumar2022gradient}. \citet{pace2024west} also consider synthetic data generation, but their purpose is to improve reward modeling in RLHF rather than sequence-level training. 
%\citet{bachmann2024pitfalls} discuss pitfalls of next token prediction from an algorithmic perspective, while we tackle the same point from a sequence-level verses token-level signal perspective.

%% file: cpo.tex
\section{Proposed approach}\label{sec:approach}
Consider a sentence of $T$ tokens $\vx=\{\vx_1,\ldots, \vx_T\}\in\gX$,
% and let $P$ be the unknown target language distribution, $\tilde P(\vx)$ be the empirical distribution of the training data (which is an approximation of $P$), and $Q$ be the distribution of our model at hand. Since our paper is also closely related to RLHF, 
 we use $\pi(\vx)$ to represent the distribution of $\vx$ under some language policy $\pi$. In particular, we write $\pi_\theta$ for a distribution that is parameterized by $\theta$, where $\theta$ is usually the set of trainable parameters of the LLM; we write $\pr$ for a reference distribution that should be clear given the context. Inspired by DPO, we introduce our CPO objective:
\begin{equation}\label{eq:norankcpo}
\resizebox{0.85\columnwidth}{!}{%
$
\begin{split}
		\mathcal{L}_{\mathrm{CPO}}\left(\pi_\theta, \pi_{\mathrm{ref}}\right)=
		\underset{\substack{ (\vx,\vy_1) \sim \mathcal{D} \\  \vy_2, \ldots, \vy_K\sim\mathcal{A}} }{\mathbb{E}} \left[\log \frac{\exp \left(\beta \log \frac{\pi_\theta\left(\vy_1 \mid \vx\right)}{\pi_{\mathrm{ref}}\left(\vy_1 \mid \vx\right)}\right)}{\sum_{j=1}^K \exp \left(\beta \log \frac{\pi_\theta\left(\vy_j \mid \vx\right)}{\pi_{\mathrm{ref}}\left(\vy_j \mid \vx\right)}\right)}\right].
\end{split}
$
}
\end{equation}
Here $(\vx, \vy_1)$ is the ground truth prefix-continuation pair from the natural language distribution $\mathcal{D}$, and $\vy_2,\ldots, \vy_K$ are $K-1$ negative continuations sampled from a to-be-discussed distribution $\mathcal{A}$. The derivation is deferred to the appendix. If some ranking of the data quality is presented, i.e. $\tau:[K]\to [K]$ where $\tau(i)<\tau(j)$ means $\vy_i$ is preferred over $\vy_j$, we also have the following CPO objective with ranking:
\begin{equation}\label{eq:rankcpo}
\resizebox{0.85\columnwidth}{!}{%
$
\begin{split}
		\mathcal{L}_{\mathrm{CPO}}\left(\pi_\theta, \pi_{\mathrm{ref}}\right)=
	\underset{\substack{\tau, (\vx,\vy_1) \sim \mathcal{D} \\  \vy_2, \ldots, \vy_K\sim\mathcal{A}} }{\mathbb{E}} \left[\log {\displaystyle \prod_{k=1}^K} \frac{\exp \left(\beta \log \frac{\pi_\theta\left(\vy_{\tau(k)} \mid \vx\right)}{\pi_{\mathrm{ref}}\left(\vy_{\tau(k)} \mid \vx\right)}\right)}{{\displaystyle \sum_{j=k}^K} \exp \left(\beta \log \frac{\pi_\theta\left(\vy_{\tau(j)} \mid \vx\right)}{\pi_{\mathrm{ref}}\left(\vy_{\tau(j)} \mid \vx\right)}\right)}\right].
\end{split}
$
}
\end{equation}

Unlike RLHF or DPO, which require human preference data $\vy_1\geq \vy_2\geq\cdots\geq\vy_K$, CPO requires only ground truth data $(x, \vy_1)\sim \mathcal{D}$, and $K-1$ synthetic negative samples $\vy_2,\ldots,\vy_K\sim\mathcal{A}$. Possibly, we can also get a ranking among the $K-1$ synthetic samples in a fully automatic way. On a high level, CPO implicitly rewards the ground truth more than the synthetic negative samples.

We consider four ways to generate synthetic data. (1) \textbf{autoregressive negatives (AN)}: We use the language model to autoregressively generate the negative samples given a prefix. We fixed the synthetic data generation strategy to be top-$k$ sampling with $k=50$. (2) $\textbf{batch negatives (BN)}$: given a batch of prefixes and continuations $\{\vx_i, \vy_i\}_{i=1}^b$, the negative samples to the prefix $\vx_i$ are composed of $\{\vy_j\}_{j\neq i}$. (3) $\textbf{meanfield negatives (MN)}$: given a sequence $\vy = \{y_1,\ldots, y_T\}$, we randomly select $c$ percent of the positions $\{t_1,\ldots, t_j\}\subseteq [T]$, and substitute each $y_{t_i}$ independently based on $\pi_\theta(y_{t_i}|y_1,\ldots,y_{t_i})$, i.e. we independently resample $c\%$ of the tokens according to their original autoregressive distribution. (4) $\textbf{truncation negatives (TN})$: for each ground truth continuation, we truncate them at a random position and append an extra EOS token at the end.
% \begin{enumerate}[noitemsep,topsep=0pt,parsep=0pt,partopsep=0pt,leftmargin=*]
% 	\item \textbf{autoregressive negatives (AN)}: We use the language model to autoregressively generate the negative samples given a prefix. We fixed the synthetic data generation strategy to be top-$k$ sampling with $k=50$.
% 	\item $\textbf{batch negatives (BN)}$: given a batch of prefixes and continuations $\{\vx_i, \vy_i\}_{i=1}^b$, the negative samples to the prefix $\vx_i$ are composed of $\{\vy_j\}_{j\neq i}$.
% 	\item $\textbf{meanfield negatives (MN)}$: given a sequence $\vy = \{y_1,\ldots, y_T\}$, we randomly select $c$ percent of the positions $\{t_1,\ldots, t_j\}\subseteq [T]$, and substitute each $y_{t_i}$ independently based on $\pi_\theta(y_{t_i}|y_1,\ldots,y_{t_i})$. 
% 	\item $\textbf{truncation negatives (TN})$: for each ground truth continuation, we truncate the continuation at a random position and append an extra EOS token at the end.
% \end{enumerate}

In our experiments, we observe that CPO can often benefit from a ranking among $K$ samples, where the ranking is based on their cosine similarity to the ground truth. Let $\ve_1,\ldots, \ve_K$ be the embeddings of given sequences $\vy_1,\ldots, \vy_K$ and without loss of generality assume that $\ve_1$ is the ground truth, we define $\tau(i)<\tau(j)$ if $\frac{\ip{\ve_i}{\ve_1}}{\|\ve_i\|\|\ve_1\|}>\frac{\ip{\ve_j}{\ve_1}}{\|\ve_j\|\|\ve_1\|}$, with the lower ranking index indicating the better sample. Using the objective \cref{eq:rankcpo}, this process gives us denser signals during training and can lead to better downstream performance.

%% file: experiment.tex
%\section{Experiment}
\section{Experimental Setup}

Throughout this section, $\textbf{BN, AN, MN, TN}$ represents batch negatives, autoregressive negatives, meanfield negatives, and truncation negatives respectively. \textbf{MixN} represents a mixed negative sampling strategy for which the details can be found in its context. We use \textbf{ANR} for models trained with autoregressive negatives and ranking signals, similarly we can denote $\textbf{MixNR}$, etc. We always randomly swap $15\%$ tokens using \textbf{MN}. Although this choice is mainly heuristic, such a ratio appears quite frequently since BERT \citep{wettig2022should}.

\vspace{-5pt}
\paragraph{Task and model.} We consider two tasks in this paper. The first is an instruction-following task, trained and evaluated on the Dolly dataset \citep{DatabricksBlog2023DollyV2}. This dataset is composed of $15011$ total instruction and response pairs. We train with $7505$ sequences and test with the rest $7506$. We use pre-trained GPT2-XL \citep{radford2019language} and OpenLlama-3B\citep{touvron2023llama,openlm2023openllama} as the base model. The second is an open-ended text generation task on Wikidump data~\citep{wikidump}. We train the OpenLlama-3B model to predict $85\%$ tokens per sample given the leading $15\%$ tokens.

\begin{figure}[tb]
\centering
	\includegraphics[width=0.9\columnwidth]{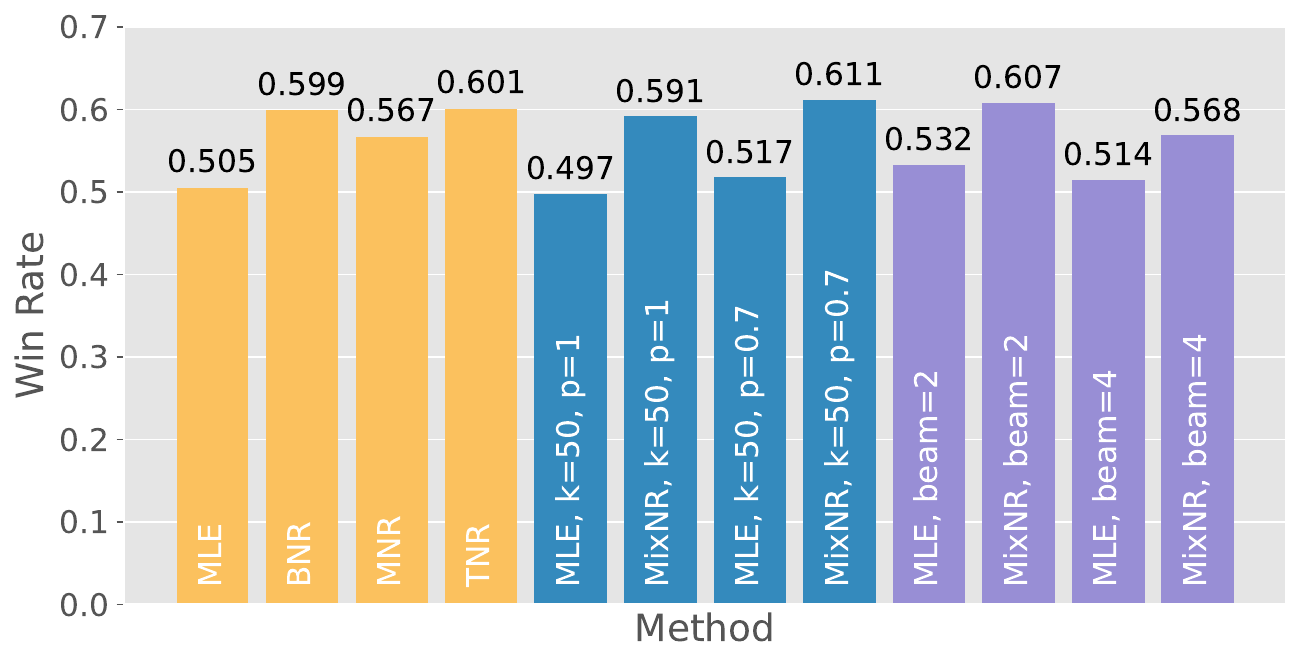}
 \vspace{-0.5em}
	\caption{The effect of different generation configuration during inference and different negative sampling methods during training. Unless otherwise specified, greedy decoding is used. Win rate is evaluated by GPT-3.5 against the ground truth continuations.}
	\label{fig:llama_comparison_neg_sample}	
\end{figure}

\vspace{-5pt}
\paragraph{Baselines.} We consider three main baselines: MLE, DPO, and parallel scheduled sampling (\textbf{PSS}, \citep{duckworth2019parallel}). Importantly, the difference between DPO and CPO lies in their negative samples. For DPO, we query GPT-3.5 to generate \textit{unhelpful} response to the Dolly instructions. PSS is trained to sample 3 sequences for each training data, and each token is replaced with $p=0.5$ (see \citet{duckworth2019parallel} for more details).
\begin{table}[tb]

 \caption{The win rate of GPT2-XL against the ground truth, samples generated by greedy decoding, evaluated by GPT-3.5.}
 \vspace{-0.5em}
\resizebox{\columnwidth}{!}{

\begin{tabular}{ c|c|c|c|c|c|c|c|c } 
  \toprule
  & \textsc{MLE} & PSS & DPO & \textsc{ANR} & \multicolumn{4}{c}{\textsc{MixNR}}  \\ 
  \midrule
  $\alpha$ & - & - & - & - & 0 & 0.5 & 0.7 & 0.9\\
  \midrule
  WinRate & 0.471 & 0.086 & 0.383 & \bf 0.506 & 0.476 & 0.479 & 0.487 & 0.485\\ 
  \bottomrule
  \hline
\end{tabular}
}
\vspace{-0.5em}
\label{table:gpteval}
\end{table}
\vspace{-5pt}
\paragraph{Training details.} Throughout the experiment, we fix the learning rate to be $1e-5$, we use the AdamW optimizer with weight decay of $0.05$. We keep the batch size to be $64$. Unless otherwise specified, for the baseline model, we train GPT2-XL and OpenLlama-3B with the next token prediction loss for $2000$ steps. Using these models as the reference model $\pr$, we continue to train with the CPO objective either with or without ranking signals, with $\beta=5$, for $1000$ steps. For both models, each training data in a batch contains $11$ negative samples in total. For MixN and MixNR, we also use a negative sample size of $11$, consisting of $3$ BN, $5$ MN, and $3$ TN. The MLE models used for evaluation are continually trained for the same number of steps from the reference model, like the CPO models. All experiments are conducted on two AWS machines, each with $8$ A100 GPUs.

\vspace{-5pt}
\paragraph{Evaluation.} As discussed in \citet{goyal2022news}, almost all automated evaluation metrics have been shown to not align with human evaluations in the modern era of LLMs, so we decide to use GPT \citep{brown2020language} as the evaluator. See the query template in the appendix.
% The query template for the Dolly instruction-following is the following: ``\texttt{For the following query to a chatbot, which response is more helpful?$\backslash$n
%     Query: \{\}$\backslash$n
%     Response A: \{\}$\backslash$n
%     Response B: \{\}$\backslash$n
%     State only "A" or "B" to indicate which response is more helpful.$\backslash$n
%     More helpful:}''
For efficiency, we generate and evaluate $1000$ samples chosen from the 7506 test set. A similar template is used for Wiki text generation, see the detail in the appendix. During inference, we consider greedy decoding, top-$p$ and top-$k$ sampling, as well as beam search.

\begin{table*}[tb]
\centering
\caption{The win rate of OpenLlama-3B trained with CPO and MLE against the ground truth data in Dolly, sampled by greedy decoding, evaluated by GPT-3.5. $ \normalfont\textsc{MLE}_1$, \textsc{ANR} and \textsc{AN} are trained for 200 steps, the rest models are trained for 1000 steps. The best CPO model outperforms the MLE baseline by 13.8\% win rate.}
\vspace{-0.5em}
\resizebox{0.9\linewidth}{!}{
\begin{tabular}{ c|c|c|c|c|c|c|c|c|c|c|c|c|c } 
  \toprule
  &$ \normalfont\textsc{MLE}_1$ & PSS & DPO & \textsc{ANR} & \textsc{AN} & $ \normalfont\textsc{MLE}_2$ & \multicolumn{6}{c|}{\textsc{MixNR}} & \textsc{MixN} \\ 
  \midrule
  $\alpha$ &  -  &   -  & - & - & - & - &0 & 0.1 & 0.3 & 0.5 & 0.7 & 0.9 & -\\
  \midrule
  WinRate &0.505 &0.270 & 0.555 & \bf 0.643 & 0.56 & 0.522 & 0.608 & 0.620 & 0.614 & 0.610 & 0.601 & 0.550 & 0.576 \\ 
  \bottomrule
  \hline
\end{tabular}
}
\vspace{-0.5em}
\label{table:openllama3beval}
\end{table*} 

\vspace{-5pt}
\paragraph{Weight-space ensemble.} Previous works \citep{liu2022brio} have also suggested to combine the auxilliary loss function with the MLE training objective $\alpha\mathcal{L}_{\mathrm{MLE}}+\mathcal{L}_{\mathrm{CPO}}$, the downside of combining loss functions in this way is that for a different choice of $\alpha$ one will have to retrain the model. To investigate the importance of loss combination, we instead perform a weight-space ensemble \citep{wortsman2022robust}. In particular, denote $\theta_{\mathrm{CPO}}$ and $\theta_{\mathrm{MLE}}$ the model parameters trained solely with CPO or MLE respectively, we generate with the interpolated weights $\theta = \alpha \theta_{\mathrm{MLE}} + (1-\alpha) \theta_{\mathrm{CPO}}$.

%\subsection{Analysis}
\section{Experimental Analysis}
\subsection{Instruction-Following Task}
Our proposed CPO method with various negative sampling strategies consistently outperforms the MLE baseline models on the Dolly instruction-following task. Using greedy sampling with GPT2-XL, the CPO model has a clear margin over the MLE model, and CPO+ANR has a $3.5\%$ higher win rate, see \cref{table:gpteval}. Note that CPO incurs very little computation overhead during the actual training: the overhead only comes a larger batch size, and even if we generate the negative samples autoregressively, it is a one-time offline cost. 

The improvement in OpenLlama-3B is more significant: CPO+ANR has a $13.8\%$ higher win rate than the MLE baseline, and CPO+MixNR has a $9.8\%$ higher win rate in \cref{table:openllama3beval}. We also observe that weight-space ensemble has a positive impact on the model. Heuristically, for OpenLlama-3B, a smaller $\alpha$ is preferred (more emphasis on the CPO weights) (\cref{table:openllama3beval}), but the reverse holds for GPT2-XL (\cref{table:gpteval}). We hypothesize that the choice of $\alpha$ should depend on the model: if the model is more capable, then it can benefit more from CPO. Here, we show the existence of a good $\alpha$, and we leave further exploration to future research. 
\vspace{-5pt}
\paragraph{Comparison with DPO and PSS.} The proposed CPO method performs better than other two baseline methods: DPO and PSS   (see Table~\ref{table:openllama3beval}).
%Interestingly, DPO does not outperform CPO (note that the difference between DPO and CPO lies in the negative samples). 
We believe that DPO performs poorly because unhelpful/irrelevant continuations (even generated by ChatGPT) do not provide a very strong signal as human generated samples. Unlike in alignment, where the toxic/harmful samples provide a clear indication of what not to generate, here it is not clear what DPO can gain from merely a single irrelevant sample. On the other hand, CPO can benefit from larger negative sample size. 
\vspace{-5pt}
\paragraph{Sampling Strategy.} In addition to greedy decoding, we also experiment with different choice of sampling strategies. In all settings, CPO has consistently demonstrated superior performance over MLE, see \cref{fig:llama_comparison_neg_sample}.

%To test whether the proposed CPO objective and the negative sampling strategies scale up to larger models, we also conduct experiments with OpenLlama-3b \citep{touvron2023llama,openlm2023openllama}. Using the same training and testing split of Dolly, we train the OpenLlama model with MLE loss for $2000$ steps as the baseline model, and continually train it with CPO for $1000$ steps. We use a negative sample size of $11$, consisting $3$ BN, $5$ MN, and $3$ TN. The greedy decoding results are listed in \cref{table:openllama3beval}. Compare to GPT2-XL, OpenLlama-3B benefits more from CPO.

\vspace{-5pt}
\paragraph{Effect of different negative samples.} We perform a study on the effects of different negative sampling strategies; the results are presented in \cref{fig:llama_comparison_neg_sample}. We first train the OpenLlama-3B model with MLE loss for 1000 steps, then continue to train with CPO for 200 steps. For all ground truth sequences, we use $4$ negative sequences. In this setting, we always use the ranking information to train CPO. We observe that the effects of BNR and TNR on the reward model preference are similar and that they perform slightly better than MNR.

\begin{table}[tb]
\centering 
\caption{OpenLlama-3B's win rate against the ground truth continuation on Wikidump. The model is trained with either MLE or CPO+BNR. Weight ensemble is adopted. The best CPO model outperforms the MLE baseline by 3\% win rate.}
\vspace{-0.5em}
\resizebox{0.9\columnwidth}{!}{

\begin{tabular}{ c|c|c|c|c|c } 
  \toprule
  & \textsc{MLE} &  \multicolumn{4}{c}{\textsc{BNR}}  \\ 
  \midrule
  $\alpha$ & - & 0 & 0.5 & 0.7 & 0.9\\
  \midrule
  WinRate & 0.508 & 0.455 & 0.505 & 0.5 & \bf 0.538\\ 
   \bottomrule
  \hline
\end{tabular}
}
\vspace{-0.5em}
\label{table:wikitexteval}
\end{table}

\subsection{Open-ended Text Generation Task}
We further test OpenLlama-3B's ability on an open-ended text generation task with CPO. Using Wikidump data \citep{wikidump}, for each test sample, we take its first $15\%$ tokens as the prefix and train the model with CPO on the rest $85\%$. For negative sampling, we use four BNR examples. The results in~\cref{table:wikitexteval} show that 
%with optimal weight interpolation coefficient $\alpha$, CPO can greatly 
CPO can improve the model's win rate against the MLE baseline by $3\%$. 
%The results also have a different pattern compared to the instruction-following task: the optimal choice of $\alpha$ shows a reverse trend. With the Dolly dataset we observes a small optimal $\alpha$, but on the Wiki dataset we see a large optimal $\alpha$. It is likely because the negative samples here are too noisy, since only $15\%$ prefixes are provided.
We observe that increasing $\alpha$ improves the score, the opposite of the instruction-following task. It is likely because the negative samples here are too noisy, since only $15\%$ prefixes are provided.

\begin{figure}
	\centering
	\includegraphics[width=0.8\columnwidth]{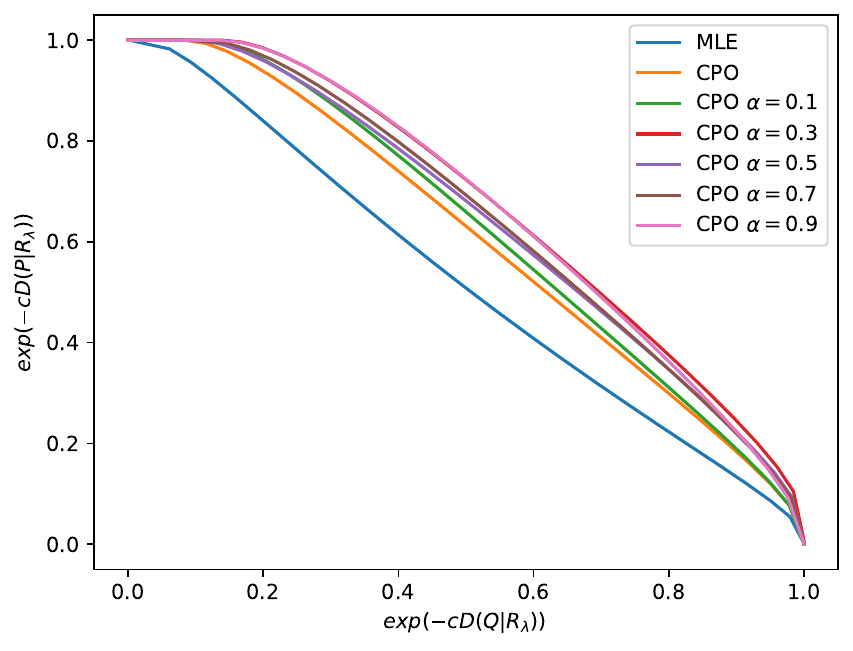}
	\caption{MAUVE score of MLE and CPO on Wiki data.}
	\label{fig:mauve}
\end{figure}

Additionally, we test the MAUVE score \citep{pillutla2021mauve} of MLE and CPO compared to the ground truth. See the results in \cref{fig:mauve} and \cref{table:wikimauve}.

\begin{table}[tb]
\centering 
\caption{MAUVE score of MLE and CPO on Wiki data.}
\vspace{-0.5em}
\resizebox{\columnwidth}{!}{
\begin{tabular}{ c|c|c|c|c|c|c|c } 
  \toprule
  & \textsc{MLE} &  \multicolumn{4}{c}{\textsc{CPO BNR}}  \\ 
  \midrule
  $\alpha$ & - & 0 & 0.1 & 0.3 & 0.5 & 0.7 & 0.9\\
  \midrule
  WinRate & 0.524 & 0.610 & 0.627 & \bf 0.673 & 0.645 & 0.651 & 0.668\\ 
   \bottomrule
  \hline
\end{tabular}
}
\vspace{-0.5em}
\label{table:wikimauve}
\end{table}

%% file: appendix.tex
\section{Appendix}

\subsection{A brief introduction to DPO, RLHF, and EBM}

\paragraph{The equivalence of RLHF and EBM} For the completeness of this paper, we include the result of the equivalence between RLHF and EBM. For the full proofs, we refer the reader to \cite{rafailov2023direct,korbak2022rl}.

The RLHF objective is the following:
\begin{align}\label{eq:rlhf_objective}
\begin{split}
		& \max_{\pi_\theta} \E_{\vx\sim\gD, \vy\sim\pi_\theta(\vy|\vx)}[r(\vx, \vy)] \\
		& \quad-\beta \KL\p{\pi_\theta(\vy|\vx)||\pr(\vy|\vx)},
\end{split}
\end{align}
where $\vx\sim\mathcal{D}$ is a given prefix, $\vy\sim\pi_\theta(\vy|\vx)$ is a sampled continuation from the trainable model $\pi_\theta$, and $r(\vx, \vy)\in \R$ is the reward. Meanwhile, we want to control the divergence between $\pi_\theta$ and $\pr$, where $\pr$ is usually an already pretrained or finetuned LLM.
The RLHF optimum is achieved at the following EBM:
\begin{align}\label{eq:rlhf_ebm}
	\pi^*(\vy|\vx) = \frac{1}{Z(\vx)}\pr(\vy|\vx)\exp\p{\frac{1}{\beta}r(\vx,\vy)},
\end{align}
where $Z(\vx)=\sum_{\vy}\pr(\vy|\vx)\exp\p{\frac{1}{\beta}r(\vx,\vy)}$ is the partition function.

\citet{rafailov2023direct} assume that the preference over two sequences $\vy_w$ and $\vy_l$ given $\vx$ is parameterized by the Bradley-Terry model:
\[
 P(\vy_w\succ\vy_l|\vx)=\frac{e^{r(\vx, \vy_w)}}{e^{r(\vx, \vy_l)}+e^{r(\vx, \vy_w)}}.
\]
% The optimal policy $\pi^*$ takes the aforementioned EBM form \cref{eq:rlhf_ebm}. and this EBM reprametrization 
Under the Bradley-Terry model, DPO establishes the equivalence between the original RLHF objective \cref{eq:rlhf_objective} and the following supervised objective:
%\begin{align}
%	\mathcal{L}_{\mathrm{DPO}}\left(\pi_\theta ; \pi_{\text {ref }}\right)=-\mathbb{E}_{\left(\vx, \vy_w, \vy_l\right) \sim \mathcal{D}}\left[\log \sigma\left(\beta \log \frac{\pi_\theta\left(\vy_w \mid \vx\right)}{\pi_{\text {ref }}\left(\vy_w \mid \vx\right)}-\beta \log \frac{\pi_\theta\left(\vy_l \mid \vx\right)}{\pi_{\text {ref }}\left(\vy_l \mid \vx\right)}\right)\right],
%\end{align}

\begin{align}
\begin{split}
	&\mathcal{L}_{\mathrm{DPO}}(\pi_\theta ; \pi_{\text{ref}})=\\
	&\mathbb{E}_{(\vx, \vy_w, \vy_l) \sim \mathcal{D}}\Big[ \log \sigma\Big(\beta \log \frac{\pi_\theta(\vy_w \mid \vx)}{\pi_{\text{ref}}(\vy_w \mid \vx)} \\
	&\quad -\beta \log \frac{\pi_\theta(\vy_l \mid \vx)}{\pi_{\text{ref}}(\vy_l \mid \vx)}\Big)\Big],
\end{split}
\end{align}

where $\sigma(\cdot)$ is the Sigmoid function.

They also generalize the formulation to the Plackett-Luce model, where we have a linear ordering $\tau(\cdot)$ among $K$ sequences:

%\begin{equation}\label{eq:rankcpo}
%\begin{split}
%		&\mathcal{L}_{\mathrm{DPO}}\left(\pi_\theta, \pi_{\mathrm{ref}}\right)\\
%		&=-\mathbb{E}_{\tau, \vy_1, \ldots, \vy_K, \vx \sim \mathcal{D}}\left[\log \prod_{k=1}^K \frac{\exp \left(\beta \log \frac{\pi_\theta\left(\vy_{\tau(k)} \mid \vx\right)}{\pi_{\mathrm{ref}}\left(\vy_{\tau(k)} \mid \vx\right)}\right)}{\sum_{j=k}^K \exp \left(\beta \log \frac{\pi_\theta\left(\vy_{\tau(j)} \mid \vx\right)}{\pi_{\mathrm{ref}}\left(\vy_{\tau(j)} \mid \vx\right)}\right)}\right].
%\end{split}
%\end{equation}

\begin{equation}\label{eq:rankdpo}
\resizebox{\columnwidth}{!}{%
$
\begin{split}
		&\mathcal{L}_{\mathrm{DPO}}\left(\pi_\theta, \pi_{\mathrm{ref}}\right)=\\
		&\underset{\substack{\tau, \vx \sim \mathcal{D} \\  \vy_1, \ldots, \vy_K} }{\mathbb{E}}\left[\log \prod_{k=1}^K \frac{\exp \left(\beta \log \frac{\pi_\theta\left(\vy_{\tau(k)} \mid \vx\right)}{\pi_{\mathrm{ref}}\left(\vy_{\tau(k)} \mid \vx\right)}\right)}{\sum_{j=k}^K \exp \left(\beta \log \frac{\pi_\theta\left(\vy_{\tau(j)} \mid \vx\right)}{\pi_{\mathrm{ref}}\left(\vy_{\tau(j)} \mid \vx\right)}\right)}\right].
\end{split}
$
}
\end{equation}
Here, $\tau(1),\ldots, \tau(K)$ induce a ranking among $K$ sequences.

\subsection{Derivation of the CPO objective function}
Here we give a full derivation of the CPO objective function in \cref{eq:norankcpo}.

Let $\vy_1,\ldots,\vy_K$ be $K$ continuations of a given prefix $\vx$. Without loss of generality, let $\vy_1$ be the best candidate. We are interested in the MLE of the event $P(\vy_1 \text{ is the best among $K$ candidates}|\vx)$.

We start from the sequence-level (RLHF) objective, notice that here $r(\cdot)$ is a reward over language quality, not human preference.
\begin{align}
\begin{split}
		& \max_{\pi_\theta} \E_{\vx\sim\gD, \vy\sim\pi_\theta(\vy|\vx)}[r(\vx, \vy)] \\
		& \quad-\beta \KL\p{\pi_\theta(\vy|\vx)||\pr(\vy|\vx)},
\end{split}
\end{align}

Its optimum is achieved at the following EBM:
\begin{align}\label{eq:rlhf_ebm_appendix}
	\pi^*(\vy|\vx) = \frac{1}{Z(\vx)}\pr(\vy|\vx)\exp\p{\frac{1}{\beta}r(\vx,\vy)},
\end{align}
where $Z(\vx)=\sum_{\vy}\pr(\vy|\vx)\exp\p{\frac{1}{\beta}r(\vx,\vy)}$ is the partition function. See the proof in \cite{rafailov2023direct,korbak2022rl}.

Now we consider the natural extension of the Bradley-Terry model to $K$ candidates:
\begin{align}\label{eq:mle_k_appendix}
\begin{split}
	&P(\vy_1 \text{ is the best among $K$ candidates}|\vx)\\
	&=\frac{\exp\p{r^*(\vx, \vy_1)}}{\sum_{k\in[K]}\exp\p{r^*(\vx, \vy_k)}}.
\end{split}
\end{align}

Now assuming we have the optimal policy $\pi^*$, we can reparameterize $r$ by rearranging \cref{eq:rlhf_ebm_appendix}:
\begin{align}\label{eq:implicit_r_appendix}
\begin{split}
	r^*(\vx, \vy)=\beta \log \frac{\pi^*(\vy \mid \vx)}{\pi_{\mathrm{ref}}(\vy \mid \vx)}+\beta \log Z(\vx).
\end{split}
\end{align}
Plugging \cref{eq:implicit_r_appendix} into \cref{eq:mle_k_appendix}, we get \cref{eq:norankcpo}.

\subsection{Query template of Dolly and Wiki text generation}
The query template for the Dolly instruction-following is the following: ``\texttt{For the following query to a chatbot, which response is more helpful?$\backslash$n
    Query: \{\}$\backslash$n
    Response A: \{\}$\backslash$n
    Response B: \{\}$\backslash$n
    State only "A" or "B" to indicate which response is more helpful.$\backslash$n
    More helpful:}''
    
The template for Wiki is the following: ``\texttt{For the following prefix, which continuation is better?$\backslash$n
    Prefix: \{\}$\backslash$n
    Continuation A: \{\}$\backslash$n
    Continuation B: \{\}$\backslash$n
    State only "A" or "B" to indicate which continuation is more helpful.$\backslash$n
    Better:}''
    
\subsection{DPO generation template and example}

When generating unhelpful responses for DPO, we query GPT with the following template:
\texttt{Given the ground truth instruction and response, can you generate a not helpful response?$\backslash$n Instruction: \{\}$\backslash$nResponse: \{\}$\backslash$nNot helpful response:}. 

One example of the generated response is the following:
\textbf{instruction:} \texttt{When did Virgin Australia start operating?}
\textbf{chosen response:} \texttt{Virgin Australia commenced services on 31 August 2000 as Virgin Blue, with two aircraft on a single route.}
\textbf{rejected response:} \texttt{Virgin Australia definitely exists and has airplanes that fly to different places.}

\subsection{Connection to noise contrastive estimation}
Noise contrastrive estimation (NCE) \citep{gutmann2010noise} is a novel estimation technique introduced to tackle the computational infeasibility of traditional likelihood-based methods in large-scale machine learning models, particularly those involving high-dimensional data. NCE diverges from typical maximum likelihood estimation by transforming the problem into a classification task, which is deeply connected to both DPO and CPO. In NCE, the model is trained to distinguish between real data and noise/synthetic data. Beyond binary classification, RankingNCE \footnote{Despite the name, it means the model is ranking the real data highest among all data, rather than learning a total ordering.} also trains the model to rank the real data higher than all noise samples \citep{ma2018noise}. 

There are two important distinctions between CPO and NCE. First, instead of training the model to distinguish between real data and noise (at which any reasonable language model should already be good), we train the model to distinguish \textit{better than a reference model does}, hence making the model better at recognizing natural text. Second, we also introduce a denser ranking signal by incorporating the similarity among embeddings of different samples. The experiments in this paper demonstrate that such a dense training signal consistently improves text generation quality.